\documentclass[review]{elsarticle}

\usepackage{lipsum}
\makeatletter
\def\ps@pprintTitle{%
 \let\@oddhead\@empty
 \let\@evenhead\@empty
 \def\@oddfoot{}%
 \let\@evenfoot\@oddfoot}
\makeatother

\usepackage{epstopdf}
\usepackage[caption=false]{subfig}

\usepackage{amsmath,xparse}
\usepackage{graphicx}
\usepackage{mathtools}

\usepackage{array}
\usepackage{textcomp}
\usepackage{gensymb}
\renewcommand{\figurename}{Fig.}

\usepackage{float}
\usepackage{placeins}

\usepackage{caption}
\usepackage{xcolor}
\usepackage{soul}
\usepackage{amstext}
\usepackage{multirow}
\usepackage{nomencl}
\usepackage{booktabs}

\usepackage{natbib}

\usepackage{ hyperref}

\begin{document}
\begin{frontmatter}

\title{A Deep Learning Framework for COVID Outbreak Prediction}

\author{Neeraj}
\ead{neeraj.pcs17@iitp.ac.in}
\author{Jimson Mathew}
\author{Ranjan Kumar Behera}
\author{Zenin Easa Panthakkalakath}

\cortext[mycorrespondingauthor]{Corresponding author at: Department of Computer Science and Engineering, Indian Institute of Technology Patna, Bihar, India.}

\address{Indian Institute of Technology Patna, Bihar, India.}

\begin{abstract}
The outbreak of COVID-19, i.e., a variation of coronavirus, also known as novel coronavirus causing respiratory disease, is a big concern worldwide since the end of December 2019. As of September 12, 2020, it has turned into an epidemic outbreak with more than 29 million confirmed cases and around 1 million reported deaths worldwide. It has created an urgent need to monitor and forecast COVID-19 spread behavior to better control this spread. Among all the popular models for COVID-19 forecasting, statistical models are receiving much attention in media. However, statistical models show less accuracy for long-term forecasting, as there is a high level of uncertainty, and the required data is also not sufficiently available. This paper proposes a comparative analysis of deep learning models to forecast the COVID-19 outbreak as an alternative to statistical models. We propose a new Attention-based encoder-decoder model named Attention-Long Short Term Memory (AttentionLSTM). LSTM based neural network layer architecture incorporates the idea of fine-grained attention mechanism, i.e., attention on hidden state dimensions instead of hidden state vector itself, highlighting the importance and contribution of each hidden state dimension. It is helpful in detection and focuses on crucial temporal information, resulting in a highly interpretable network.
Additionally, we implement a learnable vector embedding for time. As, time in a vector representation can be easily added with many architectures. This vector representation is called Time2Vec. This deep learning model forecasts the COVID-19 epidemic trend in Europe's most affected countries, i.e., Italy, Spain, France; and a North American country, i.e.,Canada.
We have used the COVID-19 data repository by the Center for Systems Science and Engineering (CSSE) at Johns Hopkins University to assess the proposed model's performance. The proposed model gives superior forecasting accuracy compared to other existing methods.
\end{abstract}

\end{frontmatter}

\section{Introduction}
The outbreak of COVID-19 first appeared in Wuhan, China, in December 2019 \cite{sohrabi2020world}. On January 30, 2020, the WHO declared a global public health emergency. Coronaviruses belong to the family Coronaviridae and are widely distributed to humans and other mammals \cite{huang2020clinical}. The infection's main symptoms are dry cough, high fever, fatigue, shortness of breath, odor, and pulmonary embolism in the most severe cases, similar symptoms induced by SARS-CoV and MERS-CoV  \cite{huang2020clinical, gralinski2020return}.
Many people may experience other symptoms, such as nausea, vomiting, and diarrhea. Some patients reported radiation changes in their ground glass lungs, normal or lower levels of white blood corpuscles, lymphocyte, platelet count, hypoxemia, liver function, and confusion. Few infected people are asymptomatic also. Most of them are said to be linked locally to Huanan's maritime market, where journalists claimed that they are selling freshly slaughtered animals. A Chinese health official said that these patients initially tested negative for viral and bacterial respiratory infections but later tested positive for coronavirus (nCoV) in the novel \cite{chan2020familial}. Initial findings suggested that the virus was not spreading person to person. It was then later confirmed in \cite{chan2020familial} that the virus spreads from person to person. It has been an epidemic of more than 15 million confirmed diseases and more than 6,000 deaths worldwide from July 22, 2020. Coronavirus was first discovered in 1965. Since then, there have been three outbreaks of the 2003 Severe Acute Respiratory Syndrome (SARS) outbreak in China \cite{gumel2004modelling, li2003angiotensin}, the 2012 outbreak of Middle East Respiratory Syndrome (MERS) in Saudi Arabia \cite{li2003angiotensin, sardar2020realistic}, and the 2015 MERS outbreak in South Korea \cite{cowling2015preliminary, kim2017middle}. The outbreak resulted in more than 8000 and 2200 confirmed cases of SARS and MERS, respectively. COVID-19 is caused by a new gene similar to the virus that causes SARS and MERS. Although the spread of COVID-19 has a lower mortality rate than the outbreak of SARS and MERS \cite{chan2020familial}, it is spreading faster and infecting more people than the outbreak of SARS and MERS \cite{chan2020familial}.

There were strict measures taken in Wuhan, China region to minimize the spread; still, the infection spread globally and turned into an international pandemic. There are differences in the number of certified cases due to differences in surveillance and detection capabilities and acquisitions between countries. Despite all efforts, however, it seems that the disease has spread worldwide today.

Since there is no cure or vaccine around to fight this virus yet, the health infrastructure and services need excellent planning. It can help in controlling the rate of disease spread. 
Thus, the total confirmed cases' estimation is essential for handling the healthcare system's demand and setting up new medical infrastructure. Various mathematical, statistical, and machine learning modeling techniques can be used to estimate short and long-term infected cases, which can help in effective planning and the number of additional materials and resources required to cope with the outbreak. This estimation of the healthcare system's anticipated burden is essential to timely and effectively manage the medical facilities and other needful resources to fight the pandemic. Such estimates can guide the severity and the number of measures needed to bring down the outbreak. In \cite{atangana2020modelling}, a new mathematical model of COVID-19 has been introduced with a lock-down effect. The authors \cite{fanelli2020analysis} analyzed and predicted COVID-19, which is widespread in China, Italy, and France. They said to find a significant drop in rising and death rates, and the infection spread should be reduced at a fast rate. \cite{torrealba2020modeling} used a mathematical model and forecasted the COVID-19 cases in Mexico. For South Africa, Turkey, and Brazil, the peak outbreak of epidemic is forecasted by analysis in \cite{djilali2020coronavirus}
In early days of pandemic, people made efforts to estimate key parameters and predict the cases in future. Statistical models \cite{lai2020assessing, chakraborty2020real} are mostly used for this. Recently, various mathematical methods such as time series models \cite{kurbalija2014time}, multivariate linear regression \cite{thomson2006potential}, and backpropagation neural networks \cite{liu2019forecasting, ren2013development, zhang2013comparative} were used to predict epidemic cases. Machine Learning (ML) has recently become famous for creating predictable models for COVID-19, as the disease spreads to be complex and large scaled in nature. ML aims at creating high-performance models with higher generalization ability and greater forecasting accuracy for longer lead-times \cite{burke2019norovirus}. Although ML methods have been used to illustrate previous epidemics (e.g. Ebola, Cholera, swine fever, H1N1 influenza, dengue fever, Zika, oyster norovirus \cite{koike2018supervised, agarwal2018data, anno2019spatiotemporal, chenar2018development, chenar2018development1, liang2020prediction, raja2019artificial, tapak2019comparative}, there is a gap in the literature for peer-reviewed paper manuscripts provided to COVID-19. Table \ref{tab1} shows the different ML and statistical models used to predict the COVID-19 epidemic. These methods are inhibited to ARIMA bases, random forest, neural networks, Bayesian networks , Naive Bayes, genetic programming and classification and regression tree (CART).
\begin{table}[!htb]
\caption{Various machine learning and statistical methods for outbreak prediction}
\label{tab1}
\centering
\begin{tabular}{|c|c|c|}
\hline
Reference  & Outbreak    & Models  \\ \hline
\cite{chen2020epidemiological} & Influenza  & SARIMA  \\ \hline
\cite{fang2020forecasting}   & Infectious Diarrhea & ARIMAX, RF \\ \hline
\cite{polwiang2020time}   & Dengue Fever  & ARIMA, ANN, MPR  \\ \hline
\cite{cao2020relationship}    & Brucellosis  & ARIMA \\ \hline
\cite{liang2020prediction}  & swine fever  & Random Forest  \\ \hline
\cite{anno2019spatiotemporal}  & Dengue fever  & ANN  \\ \hline
\cite{tapak2019comparative}   & Influenza           & Random Forest  \\ \hline
\cite{raja2019artificial}   & Dengue/Aedes        & Bayesian Network  \\ \hline
\cite{koike2018supervised}   & H1N1 Flu  & ANN    \\ \hline
\cite{agarwal2018data}  & Dengue        & \begin{tabular}[c]{@{}c@{}}Adopted multi-regression\\ and Naïve Bayes\end{tabular} \\ \hline
\cite{chenar2018development}    & Oyster norovirus    & ANN     \\ \hline
\cite{chenar2018development1}   & Oyster norovirus    & Genetic Programming  \\ \hline
\end{tabular}
\end{table}

Another class of models known as deep learning models can capture non-linear characteristics of a time series data. These models have a unique capability to capture the hidden features in the time series. 
One of the robust networks to handle sequence dependence in time-series data is Recurrent Neural Networks (RNNs). The LSTM network \cite{hochreiter1997long, yu2017long} is a special kind of RNN used in deep learning to train extensive architectures successfully. LSTMs are specially designed to handle the long-term dependency problem. The default behavior of the  LSTM network recalls information for a long period.
This paper presents a novel attention mechanism using a sequence to sequence (Seq2Seq) model. The Seq2Seq model is a deep neural network model based on LSTM units. 
The paper is organized as follows.
Sections 2 contains the essential theoretical background of methodologies used. Section 3 includes the data description and analysis. Section 4 discusses the comparative studies about the experimental results, and section 5 discusses an outlook of conclusions.

\section{Preliminaries}
\subsection{LSTM Network}
 LSTM is a unique type of RNN, having the ability to remember long-term temporal dependencies. The default property of these networks is to remember information for long periods of time. All the RNNs consist of a chain-like structure having repeating loops of the neural network. These loops help the network to retain the information in them.\par
 For an input sequence $x_1, x_2,..., x_t$, where $x_i \in {R^n}$, LSTM calculates $h_t \in {R^m}$ for each time step, t. The recurrent function of LSTM cell can be defined as follows:
 \begin{equation}
     h_t, c_t = F(h_{t-1}, c_{t-1}, x_t)
 \end{equation}
That can be defined by the following equations:

 \begin{equation}
 f_t = \sigma(W_f(h_{t-1}, x_t)+b_f)
 \end{equation}
 \begin{equation}
 i_t = \sigma(W_i.[h_{t-1}, x_t]+b_i)
 \end{equation}
 \begin{equation}
 \tilde{c} = tanh (W_c.[h_{t-1}, x_t] + b_c)
 \end{equation}
 \begin{equation}
 c_t = f_t\odot C_{t-1} + i_t \odot \tilde{c}
 \end{equation}
 \begin{equation}
o_t = \sigma(W_o.[h_{t-1}, x_t]+b_0)
 \end{equation}
 \begin{equation}
 h_t = o_t \odot tanh(c_t)
 \end{equation}

 Where $f_t, i_t, \tilde{c}, c_t$ and $o_t \in {R^{m}}$, $W_f, W_i, W_c$ and $W_o \in R^{m\times n}$, and $\odot$ represent the element-wise product.

\subsection{ Basic Attention Mechanism}
Bahdanau et al. first proposed the basic attention \cite{cho2014properties} mechanism. It calculates the weighted sum of the encoder RNN output and uses it to generate a context vector. 
Given an input ${x_1 , ... , x_T }$, it stores all the encoded data $H = {h_1 , ... , h_T }$. Here, dimension of H is $m \times n$, where m is T and n is the size of the RNN unit. The attention mechanism, which is a feed-forward neural network, accepts the previous decoder hidden state $h_t$ and one of the cell state vectors $d_{t-1}$ as input, and outputs a relevant score $e_t $. The mechanism begins with computing $e_t \in {e_1 , ... , e_T }$:
\begin{equation}
e_t = f_{att}(h_t, d_{t-1})    
\end{equation}

The attention score $\alpha_t$ , where $\alpha_t \in {\alpha_1 ,.....,\alpha_T }$, is calculated using the softmax function:
\begin{equation}
 softmax(\alpha_i) = \frac{exp(e_t)}{\sum_{t}^{ }exp(e_t))}   
\end{equation}

The context vector, $C_t \in {C_1,....., C_T }$, is the weighted sum of all the encoded data, ${h_1 , ...., h_T}$:

\begin{equation}
 C_t = \sum_{t}^{T} \alpha_t \:  . \: h_t
 \end{equation}
 
The computed $C_t$ is used to predict the output. In the training process, $C_t$ is one of of the decoder input along with  $d_{t-1}$ and $\hat{y}_{t-1}$ and it outputs $\hat{y}_{t}$. In testing process,  the output from previous step, $\hat{y}_{t-1}$, along with $d_{t-1}$ and $C_t$ , are used as the input.

\section{Data Description}
Data is collected from the COVID-19 database by the Center for Systems Science and Engineering (CSSE) at Johns Hopkins University, which can be downloaded from \url{https://github.com/CSSEGISandData/COVID-19}. The archive contains the total number of confirmed cases in the country, total deaths, and patients diagnosed, including Canada, Italy, France, and Spain, in total cases in 209 days.
\subsection{Data analysis}
\begin{figure}[!htb]
\centering
\captionsetup{justification=centering}
\includegraphics[width=8.80cm,height=5.90cm]{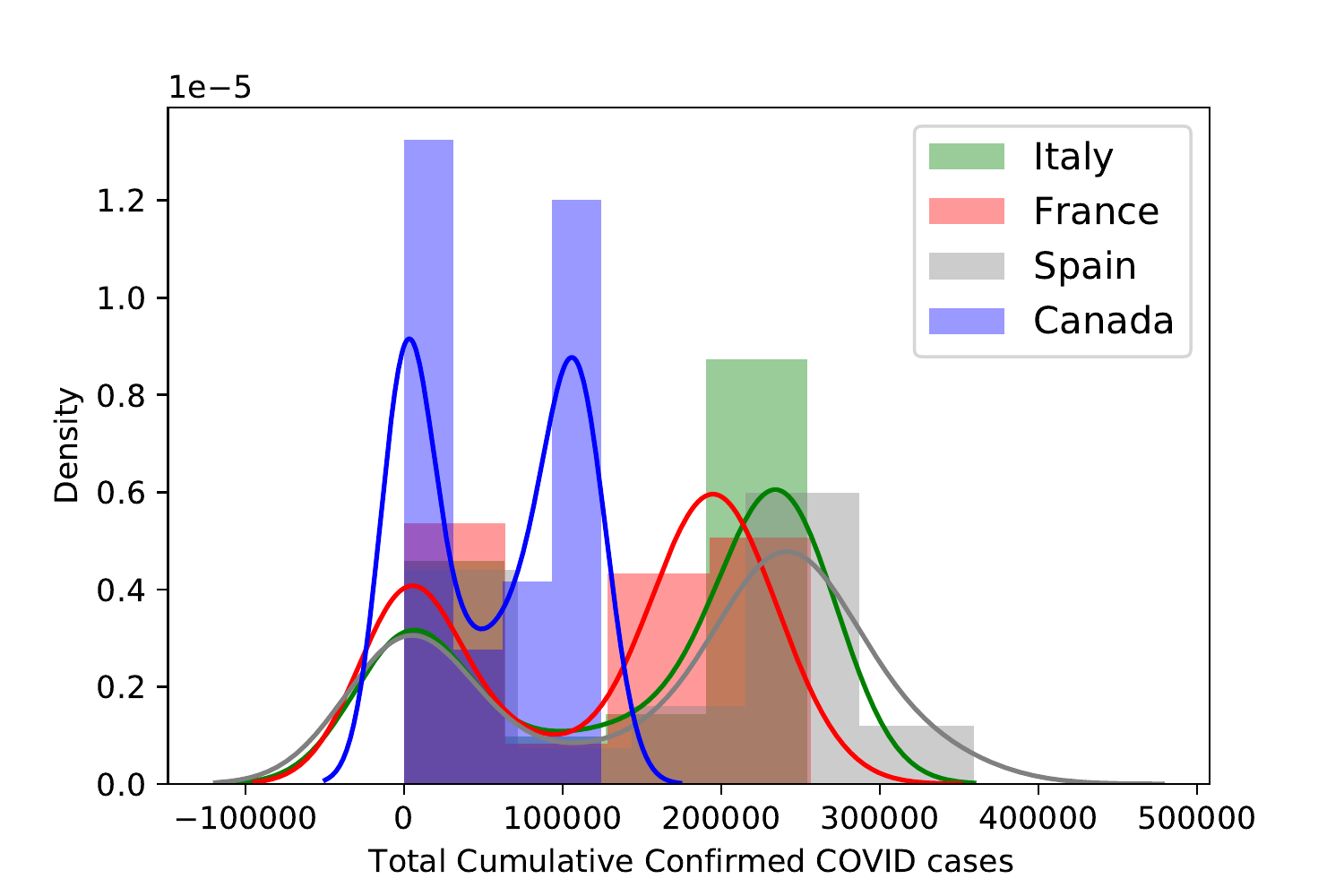}
\caption{Probability density plot for Italy, France, Spain, Canada} \label{fig1}
\end{figure}

Figure.~\ref{fig1} shows the probability density plot of all the four datasets. When we plot cumulative cases values against their probabilities, we analyze it with a random sample of a variable. By analyzing, we can determine the shape of the probability distribution, the most likely value, the spread of values, and other properties. It looks like a normal distribution. The spread of the cases is between 0 to 6 lacs. Mean, variance, kurtosis, and skewness of the distribution can be calculated with the probability distribution of a random variable. Descriptive statistics of the COVID-19 data of the considered countries between February 21, 2020, to September 12, 2020, are mentioned in Table \ref{tab2}.

\begin{figure}[!htp]
    \centering
    \begin{minipage}{0.49\textwidth}
        \centering
        \includegraphics[width=5.9cm,height=6.5cm]{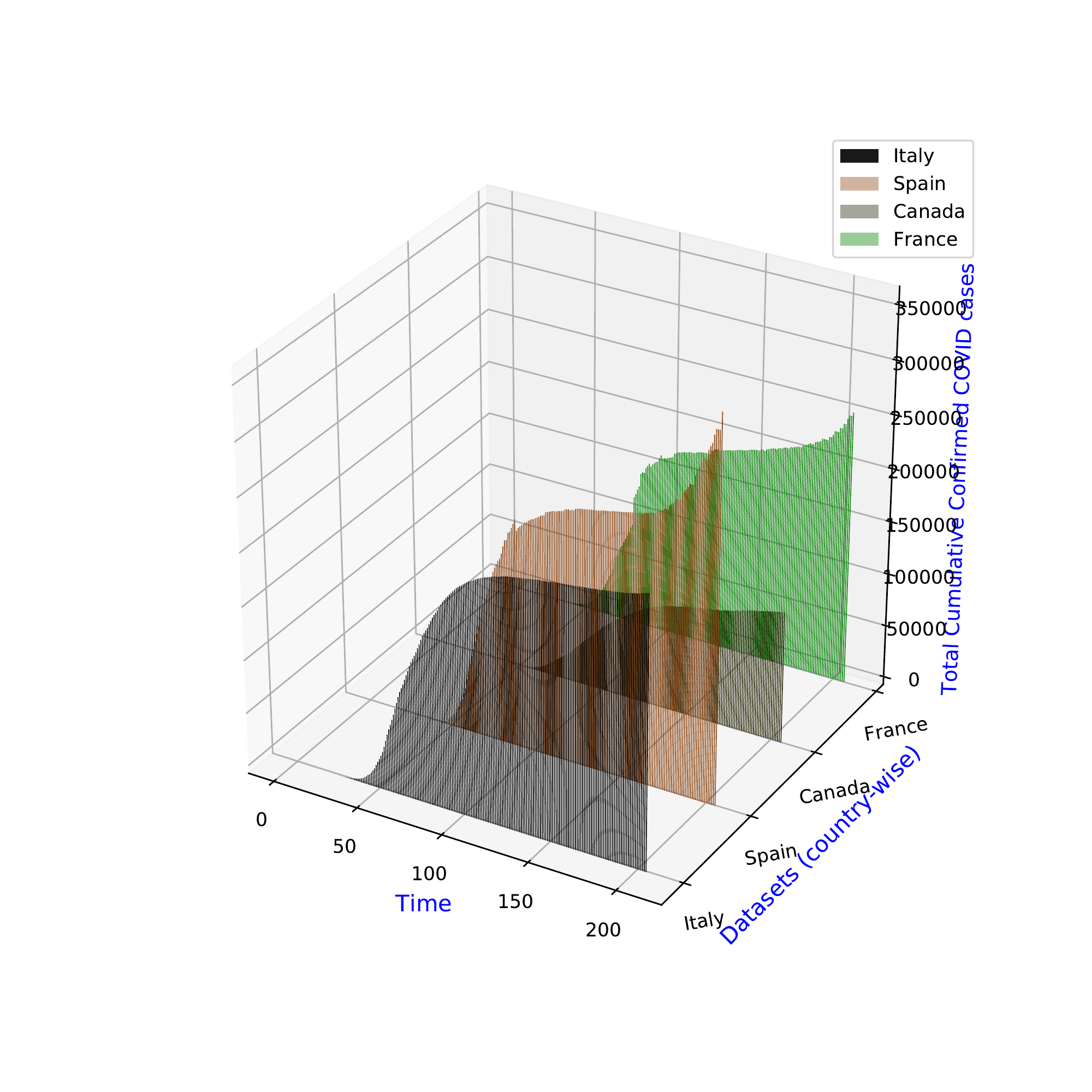} 
        \caption{3D graph showing country-wise total confirmed covid cases with time}
        \label{fig2}
    \end{minipage} \hfill
    \begin{minipage}{0.49\textwidth}
        \centering
        \includegraphics[width=5.7cm,height=6cm]{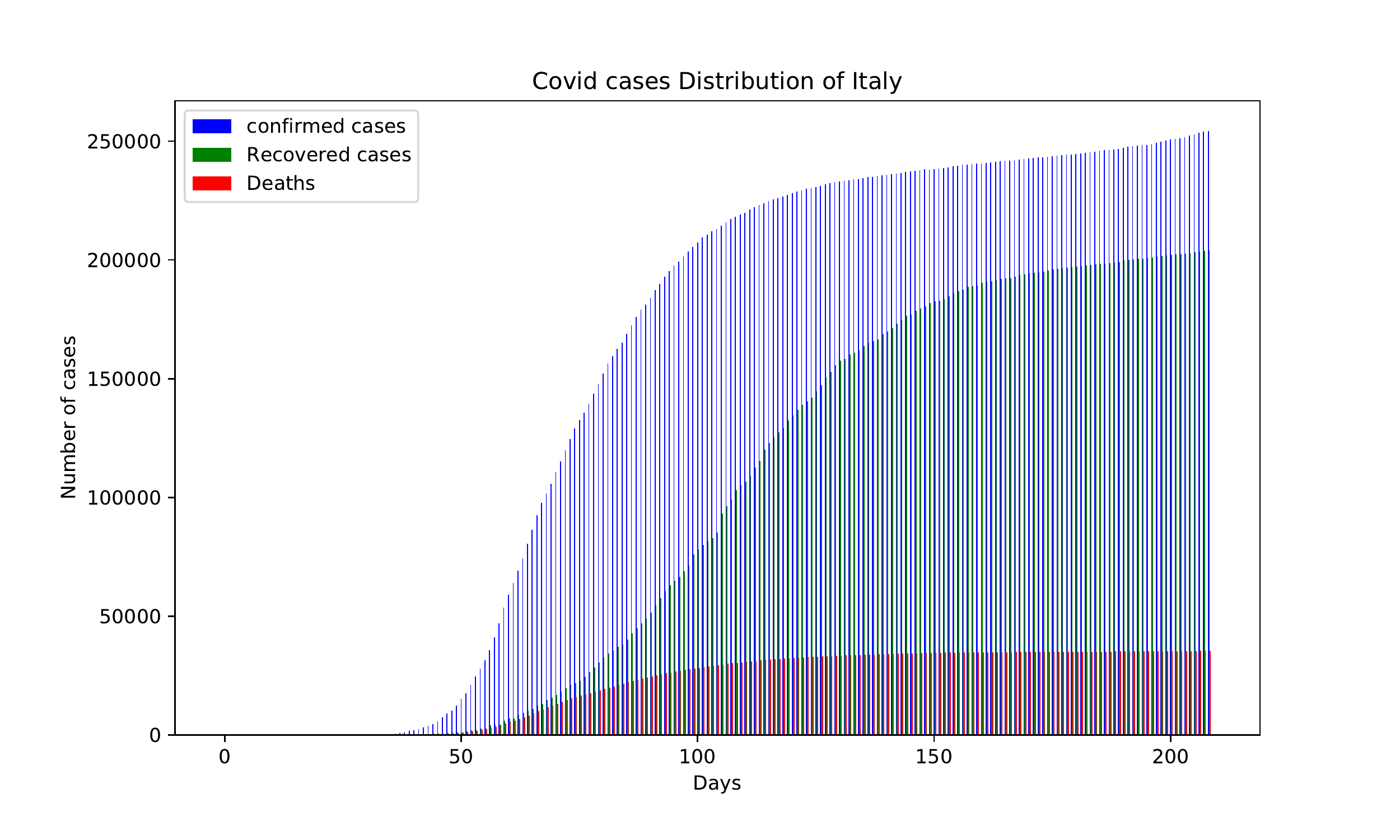} 
        \caption{Bar plot showing Total confirmed cases, recovered cases, deaths for Italy data}
        \label{fig3}
    \end{minipage}
\end{figure}

As seen in Figure.~\ref{fig2}, the outbreak of COVID-19 first started in Italy. Italy reported its first COVID-19 case on January 31, 2020. In Italy, the total number of confirmed COVID-19 cases reported during this period was 286295, with an average of 2803 new cases per day. The highest number of cases, i.e., 91,153, were reported in Lombardy, as north of Italy was affected the most by this outbreak. Neighboring Emilia-Romagna and Piedmont recorded 29,029 and 28229 cases, respectively. In terms of the number of deaths due to spread, Italy is the second largest in Europe. Spain reported the first case of COVID-19 one month after Italy, and since then, the number of confirmed cases has risen to about 650442. In France, another European country most affected, the first case of COVID-19 was reported on January 24, 2020, the death toll reached 30910, and the reported number confirmed 373911 cases.

Figure.~\ref{fig3} shows the total number of confirmed cases, recovered, and the deaths cases in Italy. From the figure, it can be seen that confirmed cases, recovered, and death cases are correlated with each other. Hence, total recovered cases and deaths are considered as supportive features while implementing the model.

\begin{table*}[!htb]
\centering
\caption{Various statistics of Italy, Spain, Canada and France data}
\label{tab2}
\resizebox{\textwidth}{!}{
\begin{tabular}{ccccccc}
\hline
Country & Mean      & Standard deviation & Minimum & Maximum  & Skewness & Kurtosis \\ \hline
Italy   & 153083.69 & 102001.56          & 0.0     & 254235.0 & -0.58    & -1.42    \\ \hline
Spain   & 162619.42 & 115449.26          & 0.0     & 359082.0 & -0.41    & -1.41    \\ \hline
Canada  & 56910.16  & 48225.29           & 0.0     & 124218.0 & -0.02    & -1.70    \\ \hline
France  & 125117.03 & 91535.22           & 0.0     & 256533.0 & -0.38    & -1.57    \\ \hline
\end{tabular}}
\end{table*}

\section{Methodology}
\subsection{Fined Grained Attention Mechanism}
The existing attention models use the single scalar score for a context vector $C_t$ at time t. It is observed that instead of using a single scalar of context vector $C_t$, it might be better to calculate and use the scalar score for each dimension of the hidden state $h_t$ at time t, as each dimension represents a different perspective into the captured internal structure. In the encoder-decoder model computation, $C_t$ shares the same attention score resulted in an equal contribution of all the dimensions of $h_t$.\par
Choi et al. \cite{choi2018fine} shows that when the different dimensions of encoded information are considered differently, and attention is applied to each dimension, it results in a better performing model.\par
 
\begin{figure}[!htb]
  \centering
  \begin{tabular}{@{}c@{}}
    \includegraphics[width=0.95\linewidth,height=65pt]{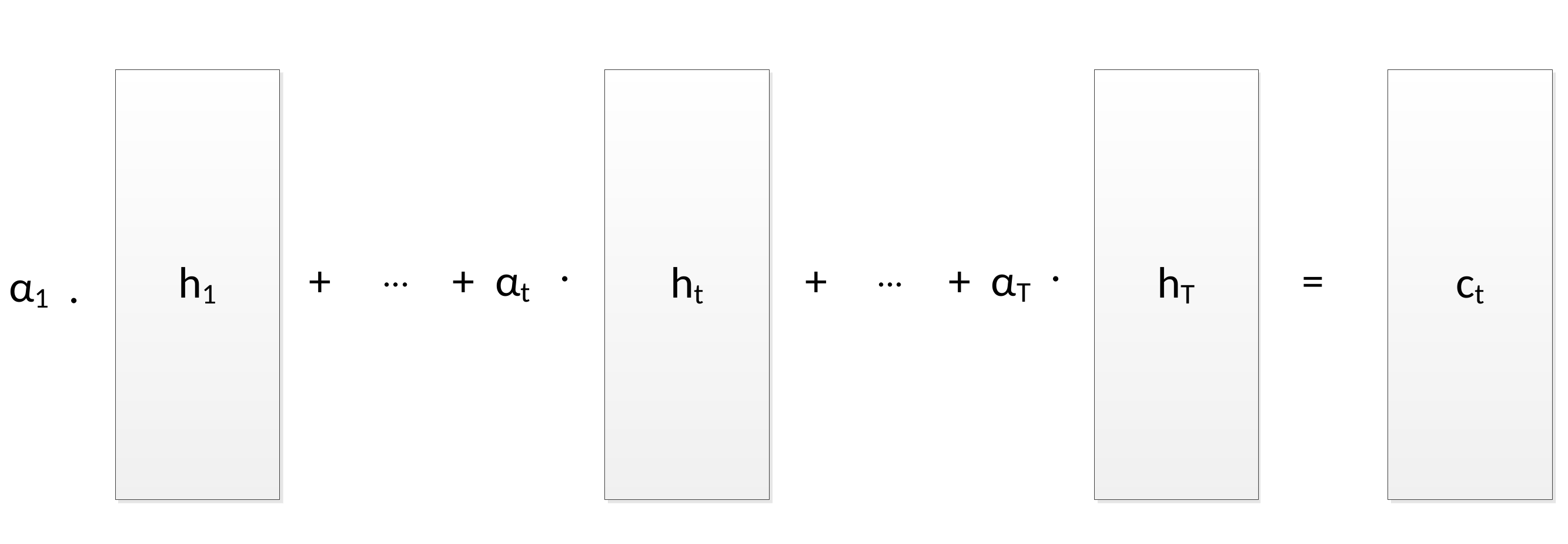} \\[\abovecaptionskip]
    \small (a) 
  \end{tabular}

  \vspace{\floatsep}

  \begin{tabular}{@{}c@{}}
    \includegraphics[width=1\linewidth,height=130pt]{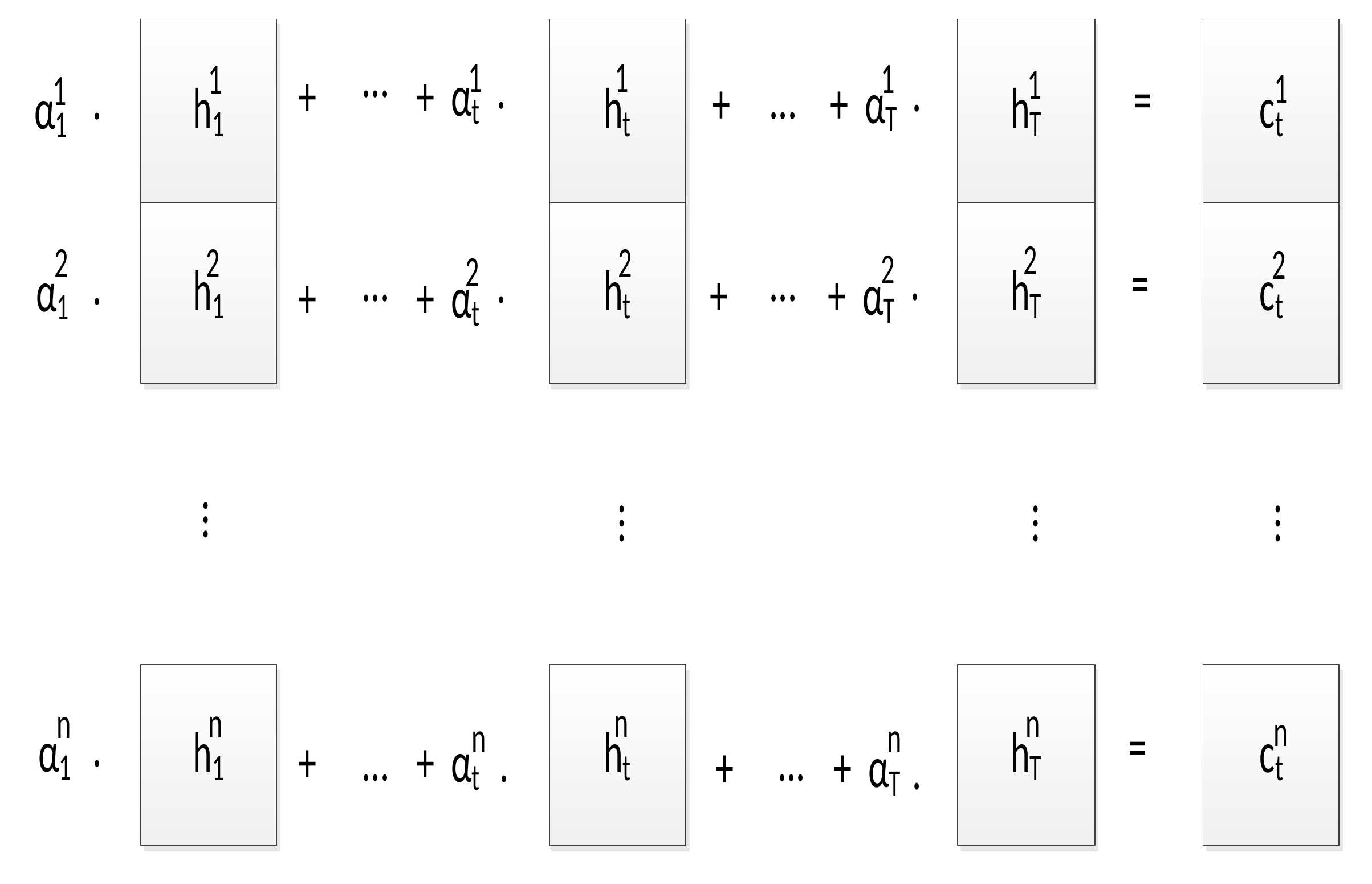} \\[\abovecaptionskip]
    \small (b)
  \end{tabular}

  \caption{(a) Basic attention process and (b) The proposed Fine grained attention mechanism}\label{fig4}
\end{figure}

Inspired by Choi et al. \cite{choi2018fine}, we propose a fine-grained attention model. 
In the proposed model, scalars are maintained for each dimension in H, which results in an increase in the number of attention scalars from $ T $ to $n \times T $.

Figure.~\ref{fig4} shows a comparison between the basic attention model and fine-grained attention model.

In the proposed function, we extend the score function $f_{Att}$ in equation \ref{12} to return a set of scores corresponding to the dimensions of the hidden state vector $h_t$ . That is,
\begin{equation}
e_t^n = f_{att}^n(h_t, d_{t-1}) 
\label{12}
\end{equation}
where $e_t^n$ is the score assigned to the $n^{th}$ dimension of the $t^{th}$ context vector $h_t$ and $f_{att}^n$ is a fully connected neural network where the number of output node is $n$. These dimension specific scores are further normalized dimension-wise such that
\begin{equation}
 \alpha_t^n = \frac{exp(e_t^n)}{\sum_{t}^{ }exp(e_t^n))}   
\end{equation}
The context vectors are then computed by
\begin{equation}
 C_t =
\begin{bmatrix} 
 \; \; \sum_{t}^{T} \alpha_t^1 \:  . \: h_t^1\; \; 
\\
\vdots 
\\
 \; \; \sum_{t}^{T} \alpha_t^n \:  . \: h_t^n\; \;  
\\ 

\end{bmatrix}
 \end{equation}

 \subsection{Time2Vec}
We used Time2Vec \cite{kazemi2019time2vec}, a representation for the time which is invariant to time rescaling.
For a given scalar notion of time t , Time2Vec of t , denoted as t2v(t), is a vector of size l + 1 defined as follows:
\begin{equation}
Time2Vec (t)[j]=\left\{\begin{matrix}
\alpha _jt + \beta _j,    \; \; \; \; if \; \; \; j = 0\\ 
G(\alpha _jt + \beta _j), \; \; \; if 1\leq j\leq l.
\end{matrix}\right.    
\end{equation}

where t2v(t)[i] is the $i^{th}$ element of t2v(t), G is a periodic activation function, and  $\alpha _j$  and $\beta _j$ are learnable parameters. Given the prevalence of vector representations for different tasks, a vector representation for time makes it easily consumable by different architectures. We chose G to be the ReLu function in our experiments.

\begin{figure}[!htb]
\centering
\captionsetup{justification=centering}
\includegraphics[width=11cm,height=7.50cm]{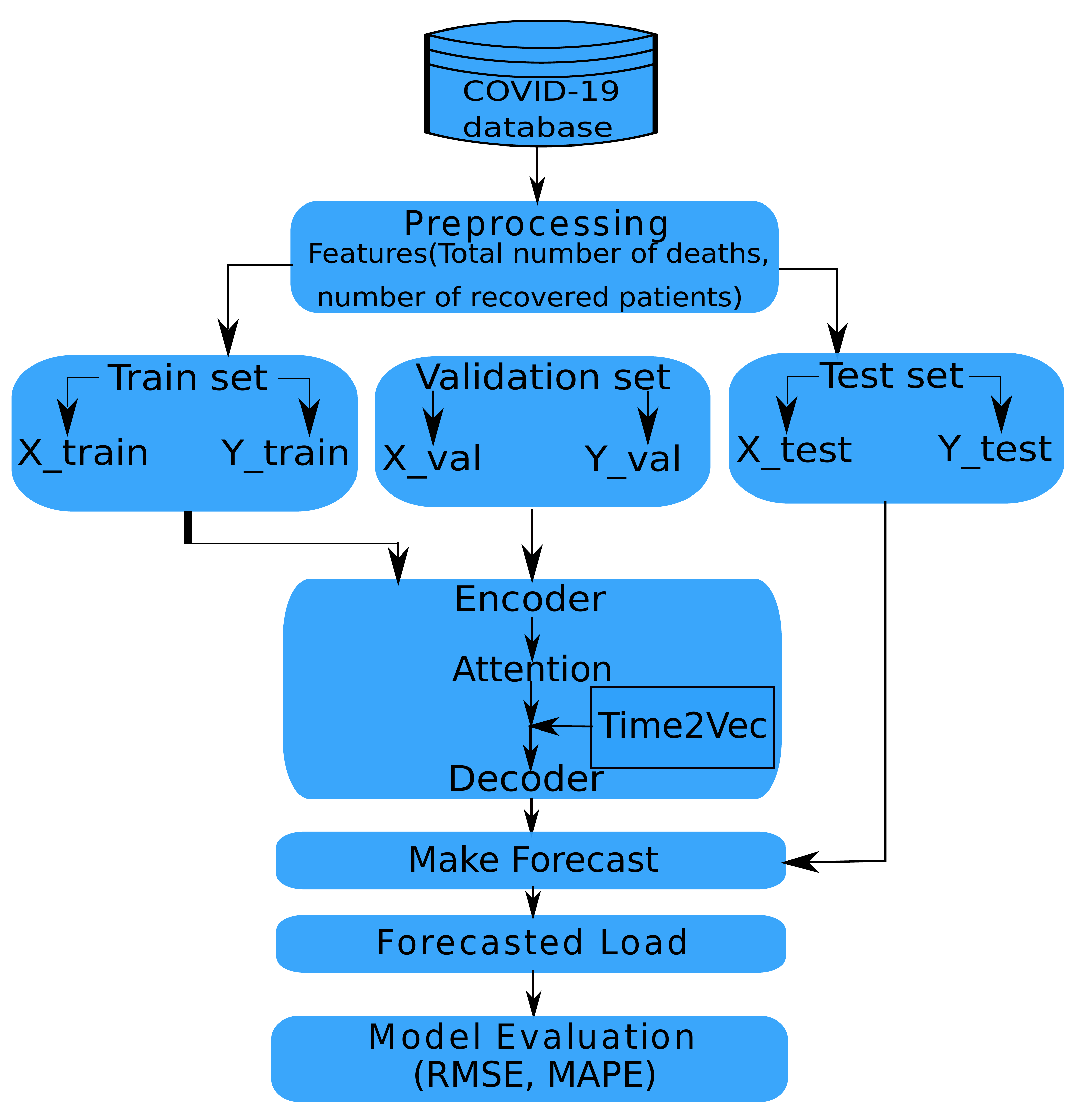}
\caption{Flow diagram of proposed methodology} \label{fig5}
\end{figure}
The flow graph of the proposed methodology is shown in Figure.~\ref{fig5}. We divided each group into training, validation and testing. The number of samples selected for training is preprocessed and scaled using a min-max scaler. Min-max scaler boosts the convergence rate of our training algorithm. The sample values are scaled between the interval  $[-1, 1]$. This also helps in preventing the large magnitude samples from bringing large weights in training.
In this paper, The LSTM based fine-grained attention model is used because it performs well for time series data. Faster convergence rate and ability to handle long-term dependencies, making them an ideal choice for the forecasting community.
The model is trained on the data on a window (lookback) basis. Lookback is the number of preceding time steps treated as input variables to predict the next period.  The predictions are made on a daily basis. We have used an exhaustive search technique for hyper-parameters tuning. The model used has an input shape of 7-time steps. Furthermore, the model has two LSTM layers, with the output sequence of each layer treated as the input for the next layer. It consists of 14 neurons in $1^{st}$, 7 neurons in $2^{nd}$ layer. We have used Rectified Linear Unit (ReLU) as the activation function for each layer. To avoid overfitting into the network, we added Batch normalization layers and Dropout with value 0.20 after each layer. A single output layer is used, which is a dense layer without an activation function. With a learning rate of 0.001 and a momentum value of 0.90, Adam optimizer is used to train the model. We trained our model for 150 epochs, and a batch size of 1 for each dataset is used. We used the Pytorch machine learning library \cite{paszke2019pytorch} to implement our model. To assess the proposed model's performance, we have used two standard error metrics: mean absolute percentage error (MAPE) and Root mean square error (RMSE).

\section{Results}
This section compared the outputs generated by the proposed method with some of the existing literature of COVID-19 outbreak prediction. The model fitted the COVID-19 data reasonably well as shown in  Table \ref{tab3} with a minimum RMSE of Italy = 689.84, RMSE of Spain = 919.27, RMSE of Canada = 36.20, and RMSE of France = 711.69 values considering 14 steps of prediction. We evaluate the model using test data and a different number of steps to predict the out of sample forecast. The number of prediction steps considered is 2, 4, 6, 8, 10, 12, and 14. The proposed model achieved better performance than other prediction models, though it is really unfortunate that transmissions are following an increasing trend. In Italy and Spain, the infection rate is increasing exponentially. The infections in Canada are also growing, though linearly, as shown in Figure \ref{fig2}.

The accuracy of all the estimates is bounded by many external factors, which puts us in a dilemma. Hence, we recommend performing further follow-up studies about the dynamic behavior of COVID-19. 
The cases reported by the Government might not be so accurate, as there is a huge backlog in getting the test results, and also, few might get immune before getting tested. These factors can also affect the accuracy of the estimation of the proposed model. The other affecting factor might be travelers traveling between different regions. Multiple trials are going on for the vaccine of COVID-19. If some of the trials successfully invent the vaccine, it might also bring down our estimates.
In Italy, the number of patients infected since February 21 follows an exponential trend until mid-May. The Government's strict restrictions helped to decline the trend, and the increase in the number of new cases was linear from mid-May to mid of August. The Government then planned to return to normal life gradually. However, it again increased the trend growth for new cases and turned into exponential growth.
Meanwhile, Spain, Europe's second-worst-hit country with 29,813 deaths, has witnessed the lower death rate between May and August. However, the death rate again started growing linearly since August.
However, the total number of confirmed cases in Spain and France has overtaken Italy. France is currently at its peak in new confirmed cases daily. Although it also observed a downward trend from May to July.  
The analysis shows that there could be a second wave of infection. This analysis can help the healthcare system to be better prepared for the pandemic.

\begin{table*}[ht!]
\centering
\caption{Generated results on Italy, Spain, Canada and France COVID-19 data}
\label{tab3}
\resizebox{\textwidth}{!}{
\begin{tabular}{cccccccccccccccccc}
\hline
\multirow{3}{*}{Dataset} & \multicolumn{1}{c|}{\multirow{3}{*}{Model}} & \multicolumn{16}{c}{Steps}                                                                                                                                                                          \\ \cline{3-18} 
                         & \multicolumn{1}{c|}{}                       & \multicolumn{2}{c}{Test} & \multicolumn{2}{c}{2} & \multicolumn{2}{c}{4} & \multicolumn{2}{c}{6} & \multicolumn{2}{c}{8} & \multicolumn{2}{c}{10} & \multicolumn{2}{c}{12} & \multicolumn{2}{c}{14} \\ \cline{3-18} 
                         & \multicolumn{1}{c|}{}                       & RMSE      & MAPE (\%)    & RMSE     & MAPE (\%)  & RMSE     & MAPE (\%)  & RMSE     & MAPE (\%)  & RMSE     & MAPE (\%)  & RMSE      & MAPE (\%)  & RMSE      & MAPE (\%)  & RMSE      & MAPE (\%)  \\ \hline
\multirow{3}{*}{Italy}   & ARIMA \cite{ceylan2020estimation}                                       & 454.66    & 2.23         & 491.34   & 2.52       & 704.53   & 4.77       & 711.98   & 5.25       & 878.34   & 6.02       & 1154.66   & 6.98       & 1251.15   & 7.53       & 1498.57   & 8.40       \\
                         & LSTM                                        & 312.10    & 2.01         & 339.09   & 2.38       & 451.34   & 4.06       & 538.41   & 4.22       & 711.05   & 6.09       & 893.25    & 7.22       & 973.78    & 6.75       & 1174.56   & 7.98       \\
                         & Proposed                                    & 209.23    & 1.71         & 217.49   & 1.86       & 576.97   & 4.21       & 479.07   & 4.12       & 606.40   & 5.96       & 678.70    & 6.02       & 692.52    & 6.18       & 689.84    & 7.07       \\
\multirow{3}{*}{Spain}   & ARIMA \cite{ceylan2020estimation}                                     & 331.12    & 2.56         & 367.72   & 2.98       & 411.03   & 3.07       & 461.21   & 3.23       & 610.73   & 4.51       & 877.40    & 5.13       & 1156.90   & 6.27       & 1389.33   & 7.94       \\
                         & LSTM                                        & 290.23    & 2.29         & 378.11   & 3.04       & 381.44   & 2.71       & 417.61   & 2.89       & 514.57   & 3.28       & 601.15    & 4.00       & 718.09    & 4.99       & 1039.90   & 7.03       \\
                         & Proposed                                    & 281.03    & 2.11         & 299.42   & 2.42       & 293.61   & 2.48       & 321.26   & 2.51       & 471.89   & 3.19       & 493.11    & 3.20       & 617.16    & 3.89       & 919.27    & 6.67       \\
\multirow{3}{*}{Canada}  & ARIMA                                       & 18.67     & 0.14         & 19.41    & 0.17       & 22.41    & 0.20       & 26.32    & 0.21       & 30.12    & 0.24       & 34.87     & 0.28       & 39.91     & 0.31       & 47.66     & 0.45       \\
                         & LSTM \cite{chimmula2020time}                                       & 13.82     & 0.12         & 15.76    & 0.14       & 19.97    & 0.16       & 22.55    & 0.20       & 26.33    & 0.22       & 32.14     & 0.25       & 37.04     & 0.29       & 46.03     & 0.43       \\
                         & Proposed                                    & 12.46     & 0.11         & 12.67    & 0.13       & 16.04    & 0.14       & 18.09    & 0.16       & 21.96    & 0.19       & 24.37     & 0.21       & 26.20     & 0.22       & 36.20     & 0.28       \\
\multirow{3}{*}{France}  & ARIMA \cite{ceylan2020estimation}                                     & 189.00    & 1.67         & 173.20   & 1.40       & 217.44   & 1.77       & 322.71   & 1.98       & 349.10   & 2.57       & 511.76    & 3.81       & 793.20    & 5.97       & 991.01    & 6.30       \\
                         & LSTM                                        & 201.49    & 1.73         & 213.77   & 1.74       & 397.01   & 4.26       & 217.79   & 1.82       & 309.14   & 2.17       & 499.71    & 3.59       & 702.83    & 5.56       & 892.74    & 6.03       \\
                         & Proposed                                    & 163.78    & 1.21         & 167.36   & 1.34       & 496.55   & 4.61       & 174.70   & 1.40       & 226.64   & 1.74       & 433.40    & 3.14       & 671.82    & 5.16       & 711.69    & 5.75       \\ \hline
\end{tabular}}
\end{table*}

\section{Conclusions}
The global pandemic of the novel coronavirus, COVID-19, has turned into many countries' primary national security issue. The development of good prediction models has become necessary to control this outbreak better and get insights into the outbreak and consequences of the infectious disease. There is a high level of uncertainty in the available data. Also, there is a lack in the availability of important data. It has affected the accuracy of standard epidemiological models, which shows low accuracy for long-term prediction. This paper presents a deep learning-based model using a fine-grained attention mechanism and vector embedding. The proposed method has been evaluated with four publicly available COVID-19 datasets.  The proposed method has been compared with ARIMA and LSTM models used in literature with various step sizes. A comparative analysis has been done to evaluate the proposed method's performance using RMSE and MAPE error matrices. The results of the proposed method reported a high generalized ability for long-term prediction. The paper suggests that the proposed model can be an effective method to model current outbreak data and predict long-term behavior.

\section{References}

 \bibliographystyle{elsarticle-num}

\bibliography{interacttfssample}
\end{document}